\documentclass[times, review, 10pt]{elsarticle}

\biboptions{numbers,sort&compress}
\usepackage{amsmath,amssymb}
\usepackage{mathrsfs}%
\usepackage{textcomp}%
\usepackage{booktabs,multirow,colortbl}
\usepackage{algorithm}%
\usepackage{algorithmicx}%
\usepackage{algpseudocode}%
\usepackage{listings}%
\usepackage{rotating}
\usepackage{adjustbox}
\usepackage{caption}
\usepackage{tabularx}
\usepackage{tikz}
\usepackage{overpic}
\usepackage{placeins}
\usepackage{xurl}
\usepackage[hidelinks]{hyperref}

\makeatletter
\let\ps@pprintTitle\ps@plain
\makeatother
	
\definecolor{first}{rgb}{1,0.75,0.75}
\definecolor{second}{rgb}{1,0.874,0.75}
\definecolor{third}{rgb}{1.,0.984,0.788}

\newdefinition{example}{Example}
\newdefinition{remark}{Remark}
\newdefinition{definition}{Definition}

\begin{document}
\let\WriteBookmarks\relax

\begin{frontmatter}

\title{Direct and Adaptable Mesh-Gaussian Scene Reconstruction from Multi-View Images}

\author[label1,fn1]{Ancheng Lin}
\author[label2,fn1]{Tianqing Su}
\author[label5]{Zuo Yuan}
\author[label3]{Quanke Su}
\author[label4]{Samuel S. Mao}
\author[label1,cor1]{Yusheng Xiang}
\ead{yusheng.xiang@ntu.edu.cn}
\ead[url]{https://orcid.org/0000-0002-4306-961X}
\cortext[cor1]{Corresponding author.}

\fntext[fn1]{These authors contributed equally to this work.}

\address[label1]{Center for Intelligent Autonomous Systems, Nantong University, Nantong 226019, Jiangsu, China}
\address[label2]{SunnyWay Tech LLC, Suzhou 215000, Jiangsu, China}
\address[label5]{Aviation Service Department, China Southern Airlines Company Limited, Guangzhou 510403, Guangdong, China}
\address[label3]{Thrust of Intelligent Transportation, The Hong Kong University of Science and Technology (Guangzhou), Guangzhou 511400, Guangdong, China}
\address[label4]{Department of Mechanical Engineering, University of California at Berkeley, Berkeley, CA 94720, USA}

\begin{abstract}
Jointly recovering explicit surface geometry and high-quality appearance from multi-view images remains challenging. This capability is essential for maintaining high-fidelity real-to-sim environments for embodied intelligence, where local changes should be incorporated without complete reconstruction. Existing neural surface reconstruction and 3DGS-to-mesh pipelines often learn geometry indirectly or separate geometry construction from appearance modeling. This separation introduces optimization redundancy and makes local geometry or appearance updates expensive. We propose an end-to-end mesh-Gaussian scene representation that binds 3D Gaussians to mesh faces and uses differentiable 3DGS rendering for photometric supervision. This design provides a direct information pathway for jointly learning explicit geometry and renderable appearance. Experiments on indoor and outdoor scenes demonstrate improved efficiency and rendering quality while preserving high-quality surface reconstruction. The explicit mesh also enables mesh-based manipulation, and the coupled representation adapts efficiently to local scene modifications. These properties support scalable visual scene modeling and the efficient maintenance of real-to-sim environments for embodied-agent training and evaluation.
\end{abstract}


\begin{keyword}
3D scene reconstruction \sep Gaussian splatting \sep Adaptable scene representation \sep Real-to-sim
\end{keyword}

\end{frontmatter}

\section{Introduction}
Real-to-sim pipelines build digital replicas of physical environments from multi-view images to train and evaluate embodied agents. They provide a practical alternative to collecting real-world interaction data, which is expensive, time-consuming, and constrained by safety and scalability. Yet these replicas are not static: objects may be moved, removed, or added as scenarios evolve. Incorporating such local changes without reconstructing the entire scene is therefore important for scalable embodied learning.

A mesh is often the preferred explicit geometry in interactive 3D environments, as it carries explicit surface information and enables efficient geometric reasoning and downstream simulation. When a mesh-based scene model must be learned and updated from visual observations, the key challenge is not only mesh extraction but also coupling geometry with appearance through a differentiable and reusable rendering pathway. Neural surface reconstruction methods \cite{Wang21_NeuS,Yariv23_BakedSDF, IJCV_LiJing2024PRSD} have been developed to represent scene geometry as an implicit signed distance function (SDF). These methods establish a connection with Neural Radiance Fields (NeRFs) \cite{Mildenhall20_NeRF}, enabling effective image-based supervision for geometry learning. However, the geometry is learned indirectly, as it involves converting the SDF to a radiance field. Additionally, NeRF-based learning requires expensive ray-marching in volumetric rendering. These factors lead to efficiency challenges in learning geometry and appearance.

\begin{figure}[tbp]
  \centering
  \includegraphics[width=0.6\columnwidth]{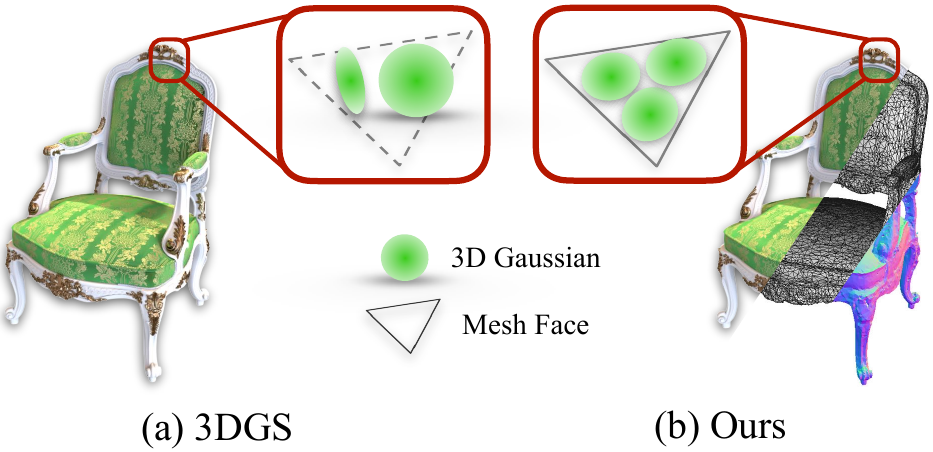}
  \caption{The Gaussians and the underlying surface do not align well in the original 3DGS \cite{Kerbl23_3DGS}, whereas our hybrid representation explicitly restricts the Gaussians to the mesh faces. Our method benefits from the high-quality rendering of 3DGS and the explicit surface structure provided by a mesh.}
  \label{fig:3dgs_vs_ours}
\end{figure}

3D Gaussian Splatting (3DGS) \cite{Kerbl23_3DGS,Zhao25_GeneralizableGS_pr} has recently gained popularity for high-quality and efficient novel-view rendering. 3DGS employs an explicit representation called \textit{Gaussians}, which represent anisotropic ellipsoids in 3D space. Gaussians enable efficient rendering due to their explicit nature and the highly parallelized splatting procedure. Despite the advantages in rendering quality and speed, it is not straightforward to reconstruct object surfaces from the traditional Gaussian representation. This is because the 3DGS model imposes no explicit constraints on the structure of the Gaussians. As illustrated in Fig. \ref{fig:3dgs_vs_ours}, Gaussians that result in visually appealing renderings may be geometrically inaccurate: they can be inconsistent with any 2D manifold. Recent methods have incorporated viewpoint and depth regularization into 3DGS optimization under sparse-view observations \cite{Zhang26_USGS_pr}. NeuSG \cite{Chen23_NeuSG} refines Gaussians using surface normals estimated by a jointly trained NeuS model \cite{Wang21_NeuS}, which notably increases the training time. SuGaR \cite{Guedon23_SuGaR} encourages Gaussians to align with the surface and subsequently converts them into a point cloud for Poisson Surface Reconstruction.

However, even when the mesh geometry itself is reconstructed from Gaussians, existing methods still rebuild the scene appearance using another independent set of Gaussian primitives, due to the superior rendering quality and efficiency of 3DGS \cite{Guedon23_SuGaR, Gao25_ManiGS, Wang25_EmbodiedGen, Sun25_RoboTidy, Qian23_GaussianAvatars, Jiang24_VRGS, Gao24_MeshBasedGS, Feng24_GS_Splash}. This two-stage pipeline incurs substantial computational overhead and optimization redundancy. Moreover, these methods are typically designed for fixed scene configurations and require running the full 3D reconstruction process whenever the environment is modified, even for minor local changes such as moving an object on a table. This limitation becomes a critical bottleneck when constructing or maintaining hundreds or thousands of scenes for embodied learning, where frequent local modifications are common. These challenges motivate us to design a more effective information pathway to supervise the learning of scene representations, while enabling efficient adaptation to local scene updates.

In this paper, we present an adaptable mesh-Gaussian scene representation that explicitly binds Gaussians to mesh faces, enabling their simultaneous learning. Specifically, the mesh is extracted from a learnable SDF of objects using differentiable marching techniques. The 3D Gaussians are then constructed as disks that are regularly attached to the mesh faces, as shown in Fig.~\ref{fig:3dgs_vs_ours}~(b). Since the 3DGS model is bound to the zero-level set of the SDF representing object surfaces, the photometric supervision applied to the 3DGS model simultaneously guides the learning of both the appearance and geometry of a scene. In addition, we model the background elements using the original 3DGS, which is jointly learned with our proposed surface-bound Gaussians. This approach makes our method more practical compared to existing methods that require inconvenient foreground masks \cite{Munkberg22_Nvdiffrec, Shen23_FlexiCubes}. As the mesh faces vary during training, the Gaussians dynamically bound to them struggle to learn per-Gaussian colors as effectively as in the original 3DGS. To address this issue, we employ a neural network to learn the scene's appearance and predict Gaussian colors for rendering.

Extensive experiments show that our method adapts efficiently to scene updates, outperforming existing two-stage methods \cite{Guedon23_SuGaR, Waczynska24_GaMeS}. It also achieves superior rendering accuracy and produces high-quality meshes in indoor and outdoor scenes, while reducing total reconstruction time by 50\% compared to recent methods and being over 10x faster than methods that rely on Neural SDFs and NeRFs \cite{Li23_Neuralangelo, Wang21_NeuS}.

In summary, our contributions are as follows:
\begin{itemize}
    \item We present a framework for directly learning an adaptable scene representation with mesh and Gaussian-based appearance in an end-to-end manner. The learned representation avoids separate mesh extraction and appearance rebuilding, enabling efficient adaptation to local scene modifications in large-scale real-to-sim environments.
    \item We propose a technique to model background appearance, extending the applicability of differentiable mesh extraction methods to outdoor scenes without the need for foreground masks.
    \item We introduce a neural appearance model that predicts Gaussian colors, enhancing the robustness of the optimization process.
\end{itemize}

\section{Related Work}
\subsection{Scene Reconstruction via Radiance Field}
Neural Radiance Field (NeRF) \cite{Mildenhall20_NeRF} is a significant advancement in 3D reconstruction. It uses neural networks to model the radiance field and yield high-fidelity images via volumetric rendering. Despite its outstanding rendering quality, NeRF requires a long time for training and rendering due to the large number of samples needed to query the neural network. Recent efforts to accelerate NeRF have employed explicit grids \cite{Keil22_Plenoxels} or hybrids of grids and small MLPs \cite{Sun22_DVGO, Muller22_iNGP, IJCV_ZhuZZZMC23}. However, these computational accelerations might compromise image quality. In addition to improving speed, some works explore scene manipulation and composition within NeRF \cite{Yang21_nerfedit}, but these are still not convenient for downstream applications. This has led to another series of works \cite{Tang23_NeRF2Mesh, Wang21_NeuS} that connect NeRF with a mesh, which will be discussed in the next section.

3D Gaussian Splatting (3DGS) \cite{Kerbl23_3DGS} employs anisotropic Gaussians to model an explicit radiance field, achieving real-time rendering and enhanced quality. The explicit nature of 3DGS makes it suitable for various domains such as scene understanding \cite{Keetha23_SplaTAM, IJCV_ZuoSZDL25} and AIGC \cite{Fang23_GaussianEditor}, where efficiency and controllability are crucial. However, directly manipulating 3DGS is challenging due to the vast number of Gaussian components involved. Hence, applications in avatar modeling \cite{Qian23_GaussianAvatars,Yin26_3DGA_pr}, VR \cite{Jiang24_VRGS}, and physics simulation \cite{Gao24_MeshBasedGS, Feng24_GS_Splash} often combine Gaussian representations with meshes or mesh-derived guidance to support manipulation and animation.

\subsection{Image-based Surface Reconstruction}
Reconstructing surfaces from images is a fundamental task in computer vision. Traditional methods, such as Multi-view Stereo (MVS) \cite{BleyerRR11_PatchMatch, Schonberger16_PixelwiseMVS, IJCV_MVS_VolumeSweeping}, estimate depth maps or extract voxel grids based on correspondences between images. However, their quality is heavily dependent on the accuracy of image matching, and the use of voxel grids is often constrained by their cubic memory requirements. Recent advancements \cite{Munkberg22_Nvdiffrec, Shen23_FlexiCubes} utilize differentiable isosurface extraction to directly optimize surface meshes under image supervision. Nonetheless, these methods are mainly for single objects due to the substantial memory demands of larger scenes.

Another promising direction, as demonstrated by NeuS \cite{Wang21_NeuS, Wang23_neus2, Li23_Neuralangelo}, adopts a hybrid representation by transforming a Signed Distance Function (SDF) into a density field, adopting the learning pipeline of NeRFs. These techniques can model view-dependent appearances and also allow for the extraction of meshes from the SDF, making them suitable for downstream applications. To enable real-time rendering, some approaches \cite{Yariv23_BakedSDF, Rakotosaona23_NeRFMeshing, Tang23_NeRF2Mesh} bake the appearance from a trained NeRF into the mesh textures. However, they typically require extensive training times and might reduce rendering quality.

Several studies have achieved mesh reconstruction using 3DGS models. DreamGaussian \cite{Tang23_DreamGaussian} evaluates the density value of grid nodes using a mixture of neighborhood Gaussians and employs the Marching Cubes algorithm \cite{Lorensen87_MarchingCubes} to extract the isosurface. However, similar to NeRFs, the resulting density field may not accurately reflect the surface, leading to unsatisfactory surface quality in complex scenes. NeuSG \cite{Chen23_NeuSG} jointly learns NeuS and 3DGS models while introducing a regularization term to ensure consistency between the Gaussian orientation and the surface normal predicted by NeuS. This joint optimization significantly extends the training duration to over ten hours. SuGaR \cite{Guedon23_SuGaR} regularizes Gaussians using derived SDF values to align them with the surface and then treats the Gaussians as a point cloud for Poisson Surface Reconstruction. However, SuGaR requires the training of an additional set of Gaussians that are bound to the extracted mesh faces. Most recently, 2DGS \cite{Huang24_2DGS} utilizes flat 2D primitives to reconstruct surfaces by fusing rendered depth and normal maps via TSDF. Similarly, GOF \cite{Yu24_GOF} enforces geometric regularization to allow for direct mesh extraction from the opacity field using Marching Cubes.

We also highlight the differences between our method and SuGaR \cite{Guedon23_SuGaR} and GaMeS \cite{Waczynska24_GaMeS}, which both construct a hybrid representation of Gaussians and mesh. Existing methods first extract the mesh or directly use the ground-truth mesh and then learn Gaussian-based appearance from scratch. In contrast, our approach simultaneously learns such a hybrid representation in an end-to-end manner. We will demonstrate that our method achieves a more efficient pipeline and better performance.

\section{Preliminary}
\subsection{3D Gaussian Splatting}

3D Gaussian Splatting (3DGS) \cite{Kerbl23_3DGS} represents a scene using anisotropic 3D Gaussians. The density of the $i$-th Gaussian is
\begin{align}
G_i(\boldsymbol{x}) = \exp\left[-\frac{1}{2}(\boldsymbol{x}-\boldsymbol{\mu}_i)^T
\boldsymbol{\Sigma}_i^{-1}(\boldsymbol{x}-\boldsymbol{\mu}_i)\right],
\end{align}
where $\boldsymbol{\mu}_i$ and $\boldsymbol{\Sigma}_i$ denote its center and covariance. Each Gaussian also has an opacity $\alpha_i$ and a view-dependent color $\boldsymbol{c}_i$ represented by Spherical Harmonics coefficients \cite{Keil22_Plenoxels}.

Given a viewing transformation $\boldsymbol{W}$, the covariance is projected into image space using the Jacobian $\boldsymbol{J}$ of the local affine approximation \cite{Zwicker01_EWA}:
\begin{align}
\boldsymbol{\Sigma}_i' = \boldsymbol{J}\boldsymbol{W}\boldsymbol{\Sigma}_i
\boldsymbol{W}^{T}\boldsymbol{J}^{T}.
\end{align}
The projected opacity at pixel $\boldsymbol{x}'$ is
\begin{align}
\alpha_i'(\boldsymbol{x}') = \alpha_i \exp\left[-\frac{1}{2}
(\boldsymbol{x}'-\boldsymbol{\mu}_i')^T\boldsymbol{\Sigma}_i'^{-1}
(\boldsymbol{x}'-\boldsymbol{\mu}_i')\right],
\end{align}
where $\boldsymbol{\mu}_i'$ is the projected center. The contributing Gaussians $\mathcal{S}$ are sorted by depth and alpha-composited as
\begin{align}
\mathbf I (\boldsymbol{x}') = \sum_{i\in\mathcal S} \boldsymbol c_i \alpha_i' \prod_{j=1}^{i-1}(1-\alpha_j'), \label{eq:alpha_blend}
\end{align}
The Gaussian parameters are optimized through photometric supervision from images with known camera poses.

\subsection{Signed Distance Function}
For a solid region $\Omega$ with boundary $\partial\Omega$, the Signed Distance Function (SDF) is
\begin{align}
D(\boldsymbol{x}) =
\begin{cases}
+d(\boldsymbol{x},\partial\Omega), & \boldsymbol{x}\notin\Omega,\\
-d(\boldsymbol{x},\partial\Omega), & \boldsymbol{x}\in\Omega,
\end{cases}
\qquad
d(\boldsymbol{x},\partial\Omega)=\min_{\boldsymbol{y}\in\partial\Omega}
\|\boldsymbol{x}-\boldsymbol{y}\|_2.
\end{align}
Its zero level set $\{\boldsymbol{x}\mid D(\boldsymbol{x})=0\}$ defines the surface geometry. Image-based methods can optimize a parameterized SDF and extract this surface using differentiable marching \cite{Wei23_NeuManifold, Shen23_FlexiCubes, Munkberg22_Nvdiffrec}. Other methods convert an SDF into a neural radiance field for image supervision \cite{Li23_Neuralangelo, Wang21_NeuS, Wang23_neus2}. Our grid parameterization and differentiable mesh extraction are described in Sec.~\ref{sec:diff_surf_extract}.

\section{Method}
\begin{figure*}[tbp]
    \centering
    \begin{overpic}[width=1.0\textwidth]{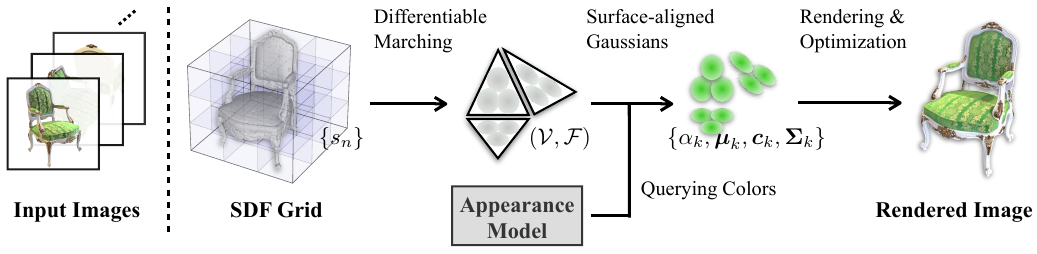}
    \fontsize{9}{5}\selectfont

    \put(36.1, 17){\color{red}\textbf{Sec.} \ref{sec:diff_surf_extract}}
    \put(56.6, 17){\color{red}\textbf{Sec.} \ref{sec:bind_gs}}
    \put(77.3, 17){\color{red}\textbf{Sec.} \ref{sec:optim}}
    \put(62, 2.9){\color{red}\textbf{Sec.} \ref{sec:gs_appearance}}
    
    \end{overpic}
    \caption{Method overview. The mesh is derived from a learnable SDF grid using a differentiable marching algorithm. Gaussians are created from the mesh faces, ensuring their alignment with the surface. A neural appearance model determines colors for Gaussians, which are then used to render an image.}
  \label{fig:overview}
\end{figure*}

The proposed method aims to obtain a set of Gaussians that strictly align with object surfaces. Specifically, these surfaces are constructed with triangle meshes, and the Gaussians are precisely defined in relation to the mesh faces. 

As illustrated in Fig. \ref{fig:overview}, our method includes modular differentiable steps, allowing the model to learn from images as in the original 3DGS. Let the 3D scene of interest be within a region $U \subset \mathbb R^3$. The model consists of:
\begin{enumerate}
    \item a signed distance function $D: U\mapsto \mathbb R$, which relates locations to the presence of objects;
    \item a triangle mesh $G=(\mathcal{V}, \mathcal{F})$ derived from $D$, with vertices $\mathcal{V} = \{\boldsymbol{v}_1, \dots, \boldsymbol{v}_V\} \subset \mathbb{R}^3$ and faces $\mathcal{F} = \{\boldsymbol{f}_1, \dots, \boldsymbol{f}_F\}$, where each $\boldsymbol{f}$ denotes a triplet of indices within $[1, \dots, V]$;
    \item a constrained 3DGS model in which the Gaussians are attached to the faces in $\mathcal{F}$;
    \item a neural appearance model, which maps a Gaussian center to spherical harmonics coefficients.
\end{enumerate}

\newcommand{\jlb}[2]{{ {\color{blue}JL:} {{\color{orange} #1}}{{\color{blue} #2}} }}

\subsection{Signed Distance Function and Differentiable Mesh Computation\label{sec:diff_surf_extract}}

We represent object surfaces using the zero level set of a signed distance function (SDF), i.e., $\{\boldsymbol{x} \mid D(\boldsymbol{x}) = 0\}$. The signed distance function $D(\boldsymbol{x})$ is defined over the entire 3D space. We represent it parametrically by specifying its values $\{s_n\}$ at the nodes $\{\boldsymbol{x}_n\}$ of a regular grid. For a generic location $\boldsymbol{x}$, $D(\boldsymbol{x})$ is computed by interpolation from the eight nodes of the grid cell enclosing $\boldsymbol{x}$.

While $D$ defines the surface implicitly, rendering and learning require explicit geometry. Marching Cubes \cite{Lorensen87_MarchingCubes} is a standard method to extract triangle meshes from an SDF. In our case, the SDF values $\{s_n\}$ are learnable and updated during training.
To enable backpropagation, we use FlexiCubes \cite{Shen23_FlexiCubes} as a differentiable mesh extractor.

Learning the SDF for outdoor and realistic scenes presents two main challenges:

\textbf{(1)} The background is defined as the complement of the foreground region $U$.
In outdoor scenes, the bounding region $\Omega$ is typically large, making the background $\Omega \setminus U$ expensive to represent jointly with $U$.
Existing methods often utilize image-space foreground segmentation masks to exclude background regions during training, thereby focusing the reconstruction process on the foreground objects \cite{Munkberg22_Nvdiffrec, Shen23_FlexiCubes}. However, this approach relies on manually annotated foreground segmentation masks, which can be labor-intensive and may not generalize well to diverse scenes. \textbf{(2)} As the scale of the foreground region $U$ increases (e.g., when $U$ occupies a substantial portion of the bounding volume $\Omega$), the cost of marching algorithms grows significantly.

To address challenge \textbf{(1)}, we introduce an additional set of 3D Gaussian splat primitives to explicitly model the visual components of the background, removing the need for manual segmentation masks. Specifically, we construct background Gaussians $\mathcal{GS}_{\text{bg}}=\{\alpha_k, \boldsymbol\mu_k, \boldsymbol c_k, \boldsymbol\Sigma_k\}$ using the original form of 3DGS. The unified representation of $\mathcal{GS}_{\text{bg}} \cup \mathcal{GS}_{\text{fg}}$ facilitates both rendering and training processes ($\mathcal{GS}_{\text{fg}}$ is detailed in Sec. \ref{sec:bind_gs}). While $\mathcal{GS}_{\text{bg}}$ introduces additional cost, it brings the practical benefit of supporting realistic novel view synthesis with background content.

For challenge \textbf{(2)}, we apply two schemes for the SDF grid to mitigate the storage and computational demands, thus enabling a larger scale of $U$:

{\bf View-Dependent Grid Optimization.}
During each training iteration, we focus on learning the scene elements that fall within a single camera's viewing frustum.
This is implemented by using the camera's intrinsic and extrinsic parameters to identify the visible grid nodes $\{\boldsymbol{x}_n\}_{n=1}^N$. 
We extract mesh faces exclusively from the grid cells composed of these visible nodes, thereby reducing computational load and memory usage.

{\bf Progressive Grid Refinement.}
In the early stages of training, the predicted SDF values are often unreliable. Initializing with a high-resolution grid tends to produce numerous mesh faces that do not correspond to meaningful surface geometry. These redundant faces increase computational overhead and disrupt gradient flow, resulting in unstable training dynamics and higher memory consumption.

To mitigate these issues, we initialize training with a low-resolution grid and progressively refine it by increasing the resolution throughout the training process. Given the node coordinates $\{\hat{\boldsymbol{x}}_n\}$ of the finer grid, their corresponding SDF values $\{\hat{s}_n\}$ are obtained by interpolating from the coarse grid. As in standard spatial queries, each $\hat{s}_n$ is interpolated from the eight nodes of the coarse grid cell enclosing $\hat{\boldsymbol{x}}_n$.

More technical details can be found in the appendix.

\begin{figure}[tbp]
  \centering
  \includegraphics[width=0.52\textwidth]{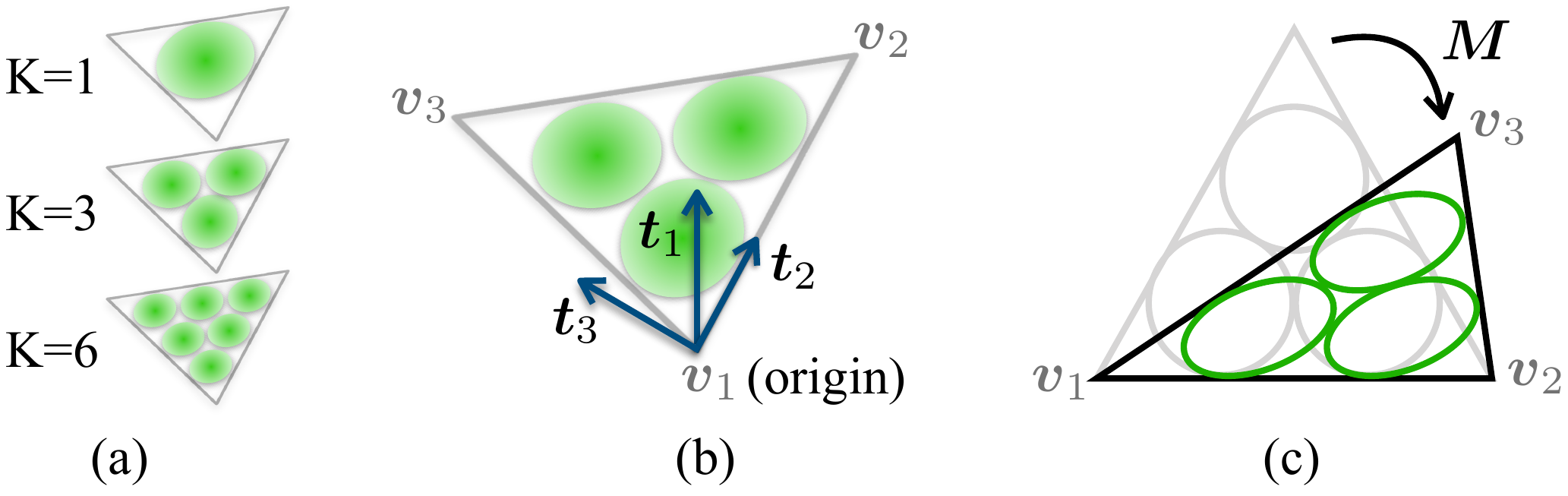}
  \caption{Subfigure (a) shows the use of $K=1,3,6$ Gaussians to represent the appearance of a triangle face, (b) defines a local coordinate frame, and (c) illustrates the application of a linear transformation $\boldsymbol{M}$ to adapt Gaussians to the irregular triangle. See Sec. \ref{sec:bind_gs} for more details.}
  \label{fig:face_space_transform}
\end{figure}

\subsection{Gaussians Consistent with Scene Geometry\label{sec:bind_gs}}

As illustrated in Fig.~\ref{fig:face_space_transform}(a), 
$K$ Gaussians are placed within each triangle.
The centers of these Gaussians are specified by barycentric coordinates:
$\boldsymbol{\xi}_1, \ldots, \boldsymbol{\xi}_K \in \mathbf{S}^2$, 
where $\mathbf{S}^2 = \{(x_1, x_2, x_3) \mid x_n \ge 0,\, x_1+x_2+x_3=1\}$ is
the standard 2-simplex.

The world-coordinate position of each Gaussian center is:
\begin{align}
\boldsymbol{\mu}_k 
= [
    \boldsymbol{v}_1, \boldsymbol{v}_2, \boldsymbol{v}_3
] \cdot \boldsymbol{\xi}_k.
\end{align}
While the barycentric coordinates $\{\boldsymbol{\xi}_k\}$ remain fixed, the
positions of the Gaussian centers $\{\boldsymbol{\mu}_k\}$ change in accordance with the mesh geometry, as they are computed based on the current positions of the mesh vertices.
Given that the mesh vertices $\boldsymbol{v}_1, \boldsymbol{v}_2,
\boldsymbol{v}_3$ are derived from the zero-level set of the learned SDF, the Gaussian centers $\{\boldsymbol{\mu}_k\}$ inherently maintain alignment with the evolving geometry during training.

During training, we compute the loss between the rendered images and ground-truth images. This loss is backpropagated through the Gaussian parameters to the mesh vertices and subsequently to the SDF values at grid nodes. This end-to-end differentiable pipeline enables the optimization of the SDF based on image supervision, ensuring consistency between the learned geometry and observed images.

To compute the Gaussian covariances, we first define a local coordinate frame as shown in Fig.~\ref{fig:face_space_transform}(b).
The frame is centered at vertex $\boldsymbol{v}_1$ (the choice of the `first vertex' is arbitrary), and its orthonormal axes are defined as $\boldsymbol{t}_i = \tilde{\boldsymbol{t}}_i / \|\tilde{\boldsymbol{t}}_i\|$, for $i = 1, 2, 3$, where
\begin{equation}
\begin{aligned}
\tilde{\boldsymbol t}_1 = \boldsymbol a \times \boldsymbol b, \quad 
\tilde{\boldsymbol t}_2 = \boldsymbol a, \quad 
\tilde{\boldsymbol t}_3 = (\boldsymbol a \times \boldsymbol b) \times
\boldsymbol a, \label{eq:triangle_coord}
\end{aligned}
\end{equation}
Here, $\boldsymbol{a} = \boldsymbol{v}_2 - \boldsymbol{v}_1$ and $\boldsymbol{b}
= \boldsymbol{v}_3 - \boldsymbol{v}_1$ are edge vectors of the triangle, and
$\times$ denotes the cross product.
The rotation matrix $\boldsymbol{R}_{t2w} = [\boldsymbol{t}_1, \boldsymbol{t}_2,
\boldsymbol{t}_3]$ maps coordinates from the local frame to the world frame.

For nearly equilateral triangles (angles close to $\pi/3$), we define the Gaussian covariance in the local frame as $\boldsymbol{\Sigma}_e = \text{diag}(\epsilon, r^2, r^2)$, modeling a thin Gaussian shell with isotropic spread over the surface patch. The radius $r$ is chosen empirically. 
For triangles with irregular aspect ratios, where the isotropic approximation is inadequate, we apply a linear transformation $\boldsymbol{M}$ to $\boldsymbol{\Sigma}_e$ (see Fig.~\ref{fig:face_space_transform}(c)). The final covariance matrix in world coordinates is given by:
\begin{align}
\boldsymbol{\Sigma} = \boldsymbol{R}_{t2w} \boldsymbol{M} \boldsymbol{\Sigma}_e
\boldsymbol{M}^T \boldsymbol{R}_{t2w}^T \label{eq:cov_final}
\end{align}
Details on the construction of $\boldsymbol{M}$ are provided in the appendix.

\subsection{Gaussian Appearance Model\label{sec:gs_appearance}}
\begin{figure}[tbp]
  \centering
  \includegraphics[width=0.6\columnwidth]{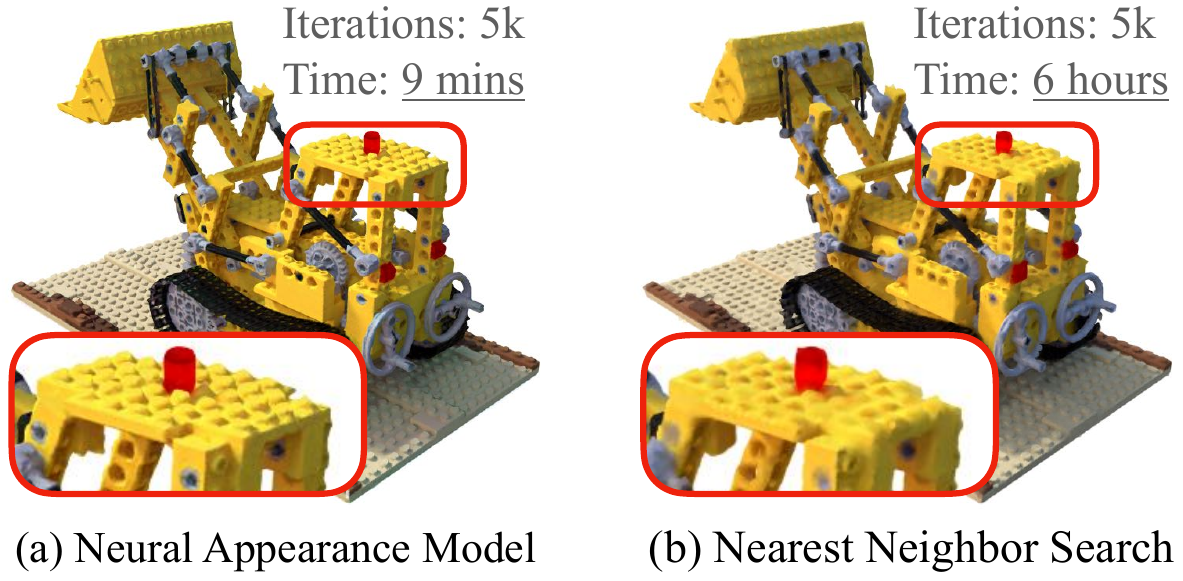}
  \caption{Comparison between two approaches to learning appearance.}
  \label{fig:appearance_neural_knn}
\end{figure}

In the original 3DGS model, each Gaussian's color $\boldsymbol{c}_k$ is directly optimized, as the set of Gaussians is relatively stable across training steps (ignoring occasional densification and pruning). In contrast, our framework dynamically generates Gaussians from the evolving SDF at each training iteration. 

While gradients from the Gaussian centers $\{\boldsymbol{\mu}_k\}$ can be
backpropagated through the shared SDF parameterization $\{s_n\}_{n=1}^N$, the
per-Gaussian colors lack such continuity, complicating their optimization. Specifically, a key challenge is the absence of explicit correspondences between
Gaussians derived from the mesh at training step $t$, denoted as
$(\mathcal{V}_t, \mathcal{F}_t)$, and those from the previous step
$(\mathcal{V}_{t-1}, \mathcal{F}_{t-1})$. A naive solution is to assign colors
to the current Gaussians by averaging the colors of neighboring Gaussians from
the previous step, defined as $\boldsymbol{c}_{i,t} = \frac{1}{\vert \mathcal{N}
\vert} \sum_{j\in \mathcal{N}} \boldsymbol{c}_{j, t-1}$, where $\mathcal{N}$ is
the set of neighbors based on the Gaussian centers. However, this method has two
major drawbacks: (1) it incurs high computational cost due to the neighbor
search between large Gaussian sets; and (2) averaging colors in local regions
can blur fine details, hindering the learning of intricate scene appearances, as
illustrated in Fig.~\ref{fig:appearance_neural_knn}(b).

To address these challenges, we introduce a neural appearance model $\mathcal{A}: \mathbb{R}^3 \rightarrow \mathbb{R}^d$, where $d$ denotes the number of Spherical Harmonics (SH) coefficients used for appearance modeling. At each training step, the SH appearance of a Gaussian centered at $\boldsymbol{\mu}$ is computed as $\boldsymbol{c} = \mathcal{A}(\boldsymbol{\mu})$. Thus, the appearance is controlled by the parameters of $\mathcal{A}$, which are shared and updated consistently throughout training.

We implement $\mathcal{A}$ using hash-grid positional encoding \cite{Muller22_iNGP} and a lightweight MLP, inspired by NvdiffRec \cite{Munkberg22_Nvdiffrec}. This design ensures that the appearance representation is compact and efficient to evaluate during training.
Since $\mathcal{A}$ predicts appearance solely based on 3D position, it remains unaffected by changes in mesh connectivity. This enables consistent learning even when the mesh changes from $(\mathcal{V}_{t-1}, \mathcal{F}_{t-1})$ to $(\mathcal{V}_t, \mathcal{F}_t)$. As shown in Fig.~\ref{fig:appearance_neural_knn}, our neural appearance model achieves nearly 40$\times$ speed-up and improved visual fidelity compared to the nearest-neighbor averaging approach.

Additionally, we fix the opacity to $\alpha = 1$ for all Gaussians to avoid semi-transparent surfaces and simplify appearance learning.

\subsection{Rendering and Optimization\label{sec:optim}}

Since our Gaussians follow the same formulation as in 3DGS \cite{Kerbl23_3DGS}, we adopt the same alpha compositing method in \eqref{eq:alpha_blend} for image rendering. For scenes with background, the rendered image includes contributions from both mesh-derived and background Gaussians, i.e., $\mathcal{GS}_{\text{bg}} \cup \mathcal{GS}_{\text{fg}}$. Our method relies solely on photometric supervision, using the loss function
\begin{align}
    \mathcal{L} = (1 - \lambda)\, \mathcal{L}_1 + \lambda\, \mathcal{L}_{\text{D-SSIM}},
\end{align}
where $\lambda$ is a weighting factor, and $\mathcal{L}_1$, $\mathcal{L}_{\text{D-SSIM}}$ denote the standard L1 and D-SSIM metrics for comparing the rendered image to the ground truth.

After the mesh and appearance model converge, we fix the topology (faces $\mathcal{F}$) and initiate a refinement stage to jointly optimize the mesh vertices and associated 3D Gaussians. In this phase, Gaussian centers remain barycentrically anchored to their corresponding triangle faces. Unlike in the earlier phase, we now treat covariance and color as per-Gaussian learnable parameters. Specifically, we initialize the SH coefficients using $\mathcal{A}$ and optimize them directly, similar to the original 3DGS. We parameterize the covariance using a 2D scaling vector $\hat{\boldsymbol{s}} \in \mathbb{R}^2$ and a complex rotation $\boldsymbol{a}i + \boldsymbol{b}$, following a scheme similar to SuGaR \cite{Guedon23_SuGaR}. This refinement step is optional, as the initial reconstruction often already yields satisfactory geometry and appearance.

\FloatBarrier

\section{Experiments}
\begin{figure*}[tbp]
  \centering
  \includegraphics[width=1.0\textwidth]{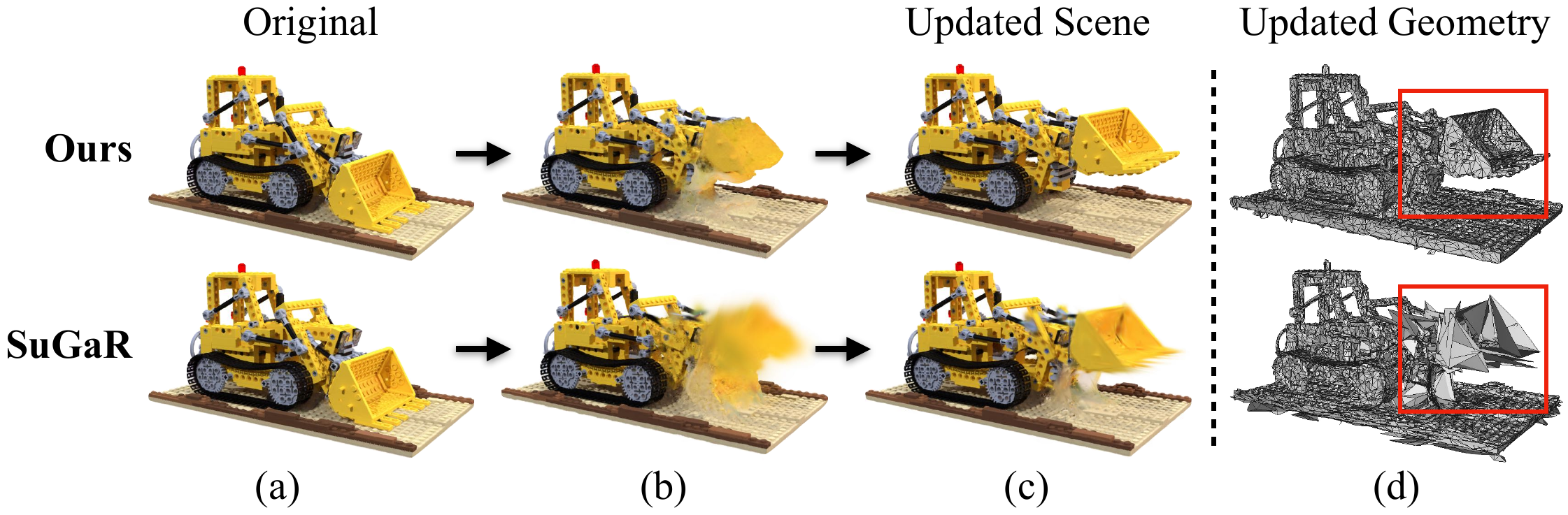}
  \caption{Adaptation to shape updates. (a) Models initially trained on the original scene.
(b) and (c) Models adapted to the updated scene for 500 and 5000 steps, respectively.
(d) The meshes learned on the updated scene.}
  \label{fig:update_scene}
\end{figure*}

\begin{figure*}[tbp]
  \centering
  \includegraphics[width=0.9\textwidth]{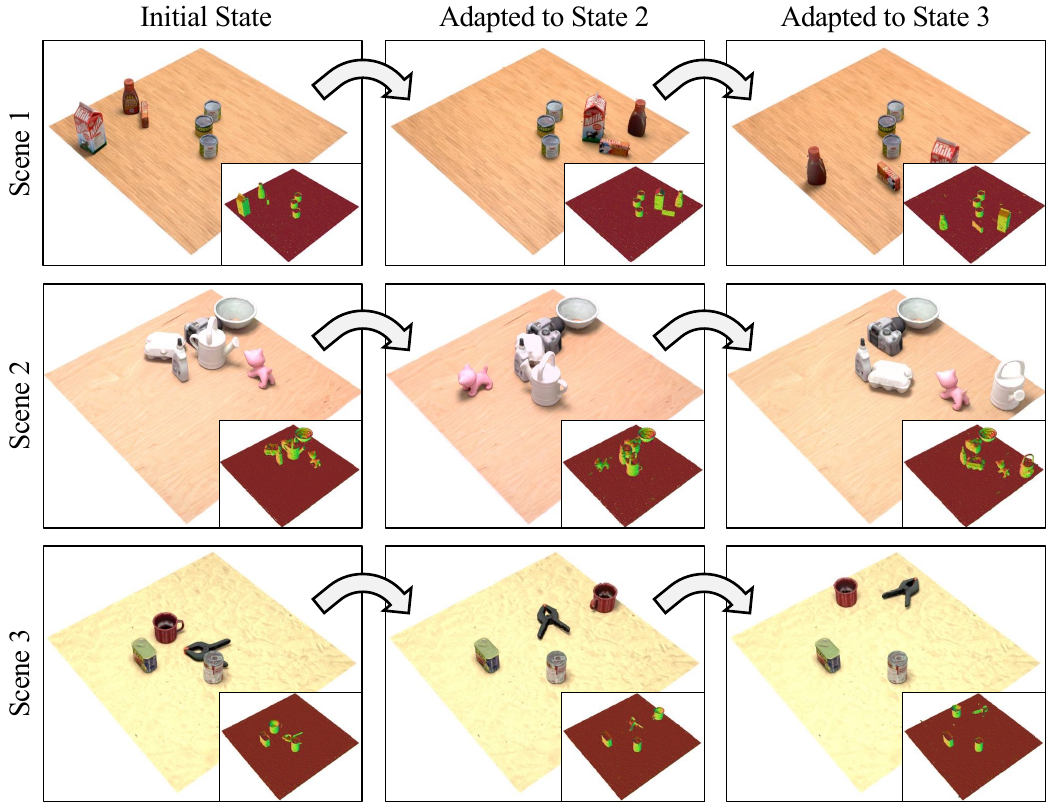}
  \caption{Adaptation to local scene modifications. For each state, we show the rendered results and the corresponding normal maps, demonstrating that our method updates both appearance and surface geometry. The adaptation converges within 5,000 optimization steps, achieving high-quality reconstruction without full retraining.}
  \label{fig:adaption_cases}
\end{figure*}

In this section, we begin by describing the implementation details across training stages, followed by the datasets used for evaluation.

The experiments are designed to evaluate our method from the perspectives of both reconstruction quality and practical applicability. In particular, we first focus on the adaptability of our representation to local scene updates, which is a key advantage over existing two-stage hybrid methods. We then evaluate rendering quality for novel view synthesis and surface reconstruction accuracy on standard benchmarks, followed by efficiency analysis and ablation studies. We compare our method with representative approaches for novel view synthesis and surface reconstruction.

Although the extracted mesh is not directly used for rendering, it serves as an explicit geometric scaffold that constrains the placement of Gaussians and enforces structural consistency. Despite this constraint, our method achieves comparable or superior performance to unconstrained methods, while enabling efficient scene manipulation.

\subsection{Implementation details}
Our method is implemented in PyTorch, using the rasterization toolkit from 3DGS \cite{Kerbl23_3DGS}. All experiments are conducted on an RTX 3090 GPU with 24~GB of memory.

\subsubsection{Initialization}
We first train the original 3DGS model for 20{,}000 iterations to obtain a set of Gaussians. Their centers are treated as a point cloud, from which we extract a coarse, watertight mesh using Alpha Shapes \cite{Edelsbrunner83_AlphaShape} and post-processing \cite{Huang18_watertight}. This mesh is then used to construct the initial SDF grid. The background Gaussians are retained and continuously optimized in subsequent stages.

Unlike methods that initialize the SDF randomly \cite{Munkberg22_Nvdiffrec,Shen23_FlexiCubes}, our strategy avoids memory overflow (OOM) issues in large-scale scenes. Thanks to the fast optimization and discrete structure of 3DGS, we can extract reliable coarse geometry within 5 to 20 minutes.

\subsubsection{Joint Learning of Mesh and Gaussians}
The SDF grid resolution is scene-dependent. For Mip-NeRF360 scenes, we use $150^3$--$250^3$ cells for the foreground, with 3 Gaussians attached per face. For NeRF-Synthetic scenes, a $100^3$ grid is used. The appearance model is implemented as an MLP with two hidden layers of 32 neurons each. This stage is optimized for 10{,}000 iterations, requiring a total of 15--30 minutes.

\subsubsection{Refinement}
Prior to refinement, we apply subdivision and decimation to standardize the mesh resolution: 0.1 million faces for single objects and 0.5 million for real-world scenes. We attach 6 Gaussians per face and initialize their covariances and colors using our adaptive construction strategy and neural appearance model. The refinement stage runs for 5{,}000--10{,}000 iterations and completes within 10--20 minutes.

\subsection{Datasets and Metrics}
The NeRF-Synthetic dataset \cite{Mildenhall20_NeRF} contains 8 scenes with full 360° coverage. In addition to evaluating novel view synthesis, we quantitatively assess the reconstructed surfaces using the provided ground-truth meshes. We also evaluate our method on 7 out of the 9 real-world scenes from the Mip-NeRF360 dataset \cite{Barron22_Mip360}, excluding \textit{Flowers} and \textit{Treehill} due to licensing constraints. We adopt the same train/test split strategy as in 3DGS \cite{Kerbl23_3DGS}.

For novel view synthesis, we use standard image quality metrics: PSNR, LPIPS, and SSIM. For surface reconstruction, we evaluate the Chamfer Distance (CD) between the extracted mesh and the ground-truth geometry.

\subsection{Adaptability to Local Scene Updates}

A key advantage of our method is the joint optimization of surface geometry and Gaussian-based appearance through their explicit binding. In contrast, existing hybrid approaches such as SuGaR \cite{Guedon23_SuGaR, Waczynska24_GaMeS, Gao24_MeshBasedGS} decouple mesh reconstruction from Gaussian optimization. This is mainly because their mesh extraction stage, which is typically based on Poisson Surface Reconstruction, is not differentiable with respect to Gaussian rasterization, resulting in a fixed surface topology during appearance learning.

Fig.~\ref{fig:update_scene} compares the results of handling local scene updates using our method and SuGaR. Our approach successfully updates both geometry and appearance in the modified regions, whereas SuGaR exhibits limited adaptability due to its fixed mesh topology, resulting in degraded reconstruction quality after scene modifications.

In practical real2sim pipelines, local scene modifications (e.g., object movement or layout changes) occur frequently. Under such conditions, our model can directly adapt to the modified regions by jointly updating geometry and appearance, without re-running the entire reconstruction process. By contrast, two-stage methods require re-extracting the mesh and retraining the Gaussian representation, leading to substantial computational overhead. Fig.~\ref{fig:adaption_cases} presents some examples, showing that our method can effectively adapt to local scene changes within only 5,000 optimization steps (approximately 5 minutes), achieving high-quality reconstruction without full retraining.

\begin{table*}[tbp]
    \centering
    \caption{Per-scene quantitative comparisons on the NeRF-Synthetic dataset \cite{Mildenhall20_NeRF}. We highlight the \colorbox{first}{best} and \colorbox{second}{second-best} results in different colors for clarity.}
    \resizebox{\linewidth}{!}{
    \begin{tabular}{lcccccccccccccccc}
    \toprule
    \multicolumn{1}{l}{\multirow{2}{1.2cm}{Method}} & \multicolumn{1}{m{0.8cm}}{\multirow{2}{1.2cm}{Surface bound}} & \multicolumn{3}{c}{Chair} & \multicolumn{3}{c}{Drums} & \multicolumn{3}{c}{Ficus} & \multicolumn{3}{c}{Hotdog}  \\
    \cmidrule(lr){3-5} \cmidrule(lr){6-8} \cmidrule(lr){9-11} \cmidrule(lr){12-14}
           & & PSNR$\uparrow$ & SSIM$\uparrow$ &LPIPS$\downarrow$ & PSNR$\uparrow$ & SSIM$\uparrow$ &LPIPS$\downarrow$ & PSNR$\uparrow$ & SSIM$\uparrow$ &LPIPS$\downarrow$ & PSNR$\uparrow$ & SSIM$\uparrow$ &LPIPS$\downarrow$ \\
    \midrule
    \textbf{without mesh} \\
    NeRF \cite{Mildenhall20_NeRF}     &   & 33.00 & 0.967 & 0.046 & 25.01 & 0.925 & 0.091 & 30.13 & 0.964 & 0.044 & 36.18 & 0.974 & 0.121 \\
    Plenoxels \cite{Keil22_Plenoxels} &   & 33.98 & 0.977 & 0.031 & 25.35 & 0.933 & 0.067 & 31.83 & 0.976 & 0.026 & 36.43 & 0.980 & 0.037 \\
    3DGS \cite{Kerbl23_3DGS} &   & 35.82 & 0.987 & 0.012 & 26.17 & 0.954 & 0.037 & 34.83 & 0.987 & 0.012 & 37.67 & 0.985 & 0.020 \\
    \midrule
    \textbf{with mesh} \\
    NVdiffRec \cite{Munkberg22_Nvdiffrec}  & $\checkmark$ & 31.60 & 0.969 & 0.034 & 24.10 & 0.916 & 0.065 & 30.88 & 0.970 & 0.041 & 33.04 & 0.973 & 0.033 \\
    NeuManifold \cite{Wei23_NeuManifold}  &  & 34.39 & 0.981 & \cellcolor{second}0.014 & 25.39 & 0.939 & 0.072 & 31.91 & 0.978 & 0.028 & 35.69 & 0.979 & 0.036 \\
    NeRF2Mesh \cite{Tang23_NeRF2Mesh}   & $\checkmark$ & 34.25 & 0.978 & 0.031 & 25.04 & 0.926 & 0.084 & 30.08 & 0.967 & 0.046 & 35.70 & 0.974 & 0.058 \\
    SuGaR-15K \cite{Guedon23_SuGaR}     & $\checkmark$  & \cellcolor{second}35.13 & \cellcolor{second}0.983 & \cellcolor{second}0.014 & \cellcolor{second}25.44 & \cellcolor{second}0.943 & \cellcolor{second}0.057 & \cellcolor{second}32.75 & \cellcolor{second}0.982 & \cellcolor{second}0.018 & \cellcolor{second}36.67 & \cellcolor{second}0.980 & \cellcolor{second}0.021 \\
    Ours                                & $\checkmark$  & \cellcolor{first}\textbf{35.40} & \cellcolor{first}\textbf{0.986} & \cellcolor{first}\textbf{0.013} & \cellcolor{first}\textbf{25.75} & \cellcolor{first}\textbf{0.952} & \cellcolor{first}\textbf{0.041} & \cellcolor{first}\textbf{34.01} & \cellcolor{first}\textbf{0.986} & \cellcolor{first}\textbf{0.014} & \cellcolor{first}\textbf{36.80} & \cellcolor{first}\textbf{0.985} & \cellcolor{first}\textbf{0.018} \\
    \toprule
    \multicolumn{1}{l}{\multirow{2}{1.2cm}{Method}} & 
    \multicolumn{1}{m{0.8cm}}{\multirow{2}{1.2cm}{Surface bound}} & \multicolumn{3}{c}{Lego} & \multicolumn{3}{c}{Materials} & \multicolumn{3}{c}{Mic}  & \multicolumn{3}{c}{Ship}  \\
    \cmidrule(lr){3-5} \cmidrule(lr){6-8} \cmidrule(lr){9-11} \cmidrule(lr){12-14}
           & & PSNR$\uparrow$ & SSIM$\uparrow$ &LPIPS$\downarrow$ & PSNR$\uparrow$ & SSIM$\uparrow$ &LPIPS$\downarrow$ & PSNR$\uparrow$ & SSIM$\uparrow$ &LPIPS$\downarrow$ & PSNR$\uparrow$ & SSIM$\uparrow$ &LPIPS$\downarrow$ \\
    \midrule
    \textbf{without mesh} \\
    NeRF \cite{Mildenhall20_NeRF}    & & 32.54 & 0.961 & 0.050 & 29.62 & 0.949 & 0.063 & 32.91 & 0.980 & 0.028 & 28.65 & 0.856 & 0.206 \\
    Plenoxels \cite{Keil22_Plenoxels} & & 34.10 & 0.975 & 0.028 & 29.14 & 0.949 & 0.057 & 33.26 & 0.985 & 0.015 & 29.62 & 0.890 & 0.134 \\
    3DGS \cite{Kerbl23_3DGS} & & 35.69 & 0.983 & 0.016 & 30.00 & 0.960 & 0.034 & 35.34 & 0.991 & 0.006 & 30.87 & 0.907 & 0.106 \\
    \midrule
    \textbf{with mesh} \\
    NVdiffRec \cite{Munkberg22_Nvdiffrec} & $\checkmark$ & 29.14 & 0.949 & 0.042 & 26.74 & 0.923 & 0.060 & 30.78 & 0.977 & 0.024 & 26.12 & 0.833 & \cellcolor{first}\textbf{0.080} \\
    NeuManifold \cite{Wei23_NeuManifold}  & & 34.00 & 0.977 & 0.024 & 26.69 & 0.924 & 0.115 & 33.40 & 0.986 & 0.012 & 28.63 & 0.875 & 0.168 \\
    NeRF2Mesh \cite{Tang23_NeRF2Mesh}     & $\checkmark$ & \cellcolor{second}34.90 & 0.977 & 0.025 & 26.26 & 0.906 & 0.111 & 32.63 & 0.979 & 0.038 & \cellcolor{first}\textbf{29.47} & 0.875 & 0.138 \\
    SuGaR-15K \cite{Guedon23_SuGaR}     & $\checkmark$  & 34.87 & \cellcolor{second}0.981 & \cellcolor{second}0.015 & \cellcolor{second}27.86 & \cellcolor{second}0.944 & \cellcolor{second}0.046 & \cellcolor{second}34.81 & \cellcolor{second}0.989 & \cellcolor{second}0.008 & \cellcolor{second}29.22 & \cellcolor{second}0.880 & \cellcolor{second}0.103 \\
    Ours                                 & $\checkmark$   & \cellcolor{first}\textbf{34.94} & \cellcolor{first}\textbf{0.982} & \cellcolor{first}\textbf{0.016} & \cellcolor{first}\textbf{27.93} & \cellcolor{first}\textbf{0.950} & \cellcolor{first}\textbf{0.043} & \cellcolor{first}\textbf{34.89} & \cellcolor{first}\textbf{0.991} & \cellcolor{first}\textbf{0.007} & 28.32 & \cellcolor{first}\textbf{0.883} & 0.121 \\
    \bottomrule
    \end{tabular}
    \label{tab:NVS_syn}
    }
    
\end{table*}

\begin{table*}[tbp]
    \centering
    \caption{Quantitative comparisons on the Mip-NeRF360 dataset \cite{Barron22_Mip360}. We highlight the \colorbox{first}{best}, \colorbox{second}{second-best}, and \colorbox{third}{third-best} results in different colors for clarity.}
    \resizebox{\linewidth}{!}{
    \begin{tabular}{lcccccccccccc}
    \toprule
    \multicolumn{1}{l}{\multirow{2}{1.2cm}{Method}} &
    \multicolumn{1}{p{0.8cm}}{\multirow{2}{1.2cm}{Surface bound}} & \multicolumn{3}{c}{Indoor scenes} & \multicolumn{3}{c}{Outdoor scenes} & \multicolumn{3}{c}{All scenes}  \\
    \cmidrule(lr){3-5} \cmidrule(lr){6-8} \cmidrule(lr){9-11}
           & & PSNR$\uparrow$ & SSIM$\uparrow$ &LPIPS$\downarrow$ & PSNR$\uparrow$ & SSIM$\uparrow$ &LPIPS$\downarrow$ & PSNR$\uparrow$ & SSIM$\uparrow$ &LPIPS$\downarrow$ \\
    \midrule
    \textbf{without mesh} \\
    Plenoxels \cite{Keil22_Plenoxels}     &  & 24.83 & 0.766 & 0.426 & 22.02 & 0.542 & 0.465 & 23.62 & 0.670 & 0.443 \\
    INGP-Base \cite{Muller22_iNGP}       &   & 28.65 & 0.840 & 0.281 & 23.47 & 0.571 & 0.416 & 26.43 & 0.725 & 0.339 \\
    3DGS \cite{Kerbl23_3DGS}             &   & 30.41 & 0.920 & 0.189 & 26.40 & 0.805 & 0.173 & 28.69 & 0.870 & 0.182 \\
    \midrule
    \textbf{with mesh} \\
    BakedSDF \cite{Yariv23_BakedSDF}      &  & 27.06 & 0.836 & 0.258 &   -   &   -   &   -   &   -   &   -   &   -   \\
    NeuManifold \cite{Wei23_NeuManifold}  &   & 27.16 & 0.813 & 0.316 &   -   &   -   &   -   &   -   &   -   &   -   \\
    NeRF2Mesh \cite{Tang23_NeRF2Mesh}    & $\checkmark$ &   -   &   -   &   -   & 22.74 & 0.523 & 0.457 &   -   &   -   &   -   \\
    SuGaR-15K \cite{Guedon23_SuGaR}      & $\checkmark$   & 29.43 & 0.910 & 0.216 & 24.40 & \cellcolor{third}0.699 & 0.301 & 27.27 & 0.820 & 0.253 \\
    2DGS \cite{Huang24_2DGS}             &    & \cellcolor{second}30.40 & \cellcolor{second}0.916 & \cellcolor{third}0.195 & \cellcolor{third}24.34 & 0.717 & \cellcolor{third}0.246 & - & - & - \\
    GOF \cite{Yu24_GOF}                  &    & \cellcolor{first}\textbf{30.79} & \cellcolor{first}\textbf{0.924} & \cellcolor{first}\textbf{0.184} & \cellcolor{second}24.82 & \cellcolor{second}0.750 & \cellcolor{first}\textbf{0.202} & - & - & - \\
    Ours                                 & $\checkmark$   & \cellcolor{third}29.33 & \cellcolor{third}0.910 & \cellcolor{second}0.193 
                                         & \cellcolor{first}\textbf{25.17} & \cellcolor{first}\textbf{0.754} & \cellcolor{second}0.237 
    & 27.54 & 0.843 & 0.212 \\
    \midrule
    \end{tabular}
    \label{tab:NVS_mip}
    }
    
\end{table*}
\begin{figure*}[tbp]
  \centering
  \includegraphics[width=1.0\textwidth]{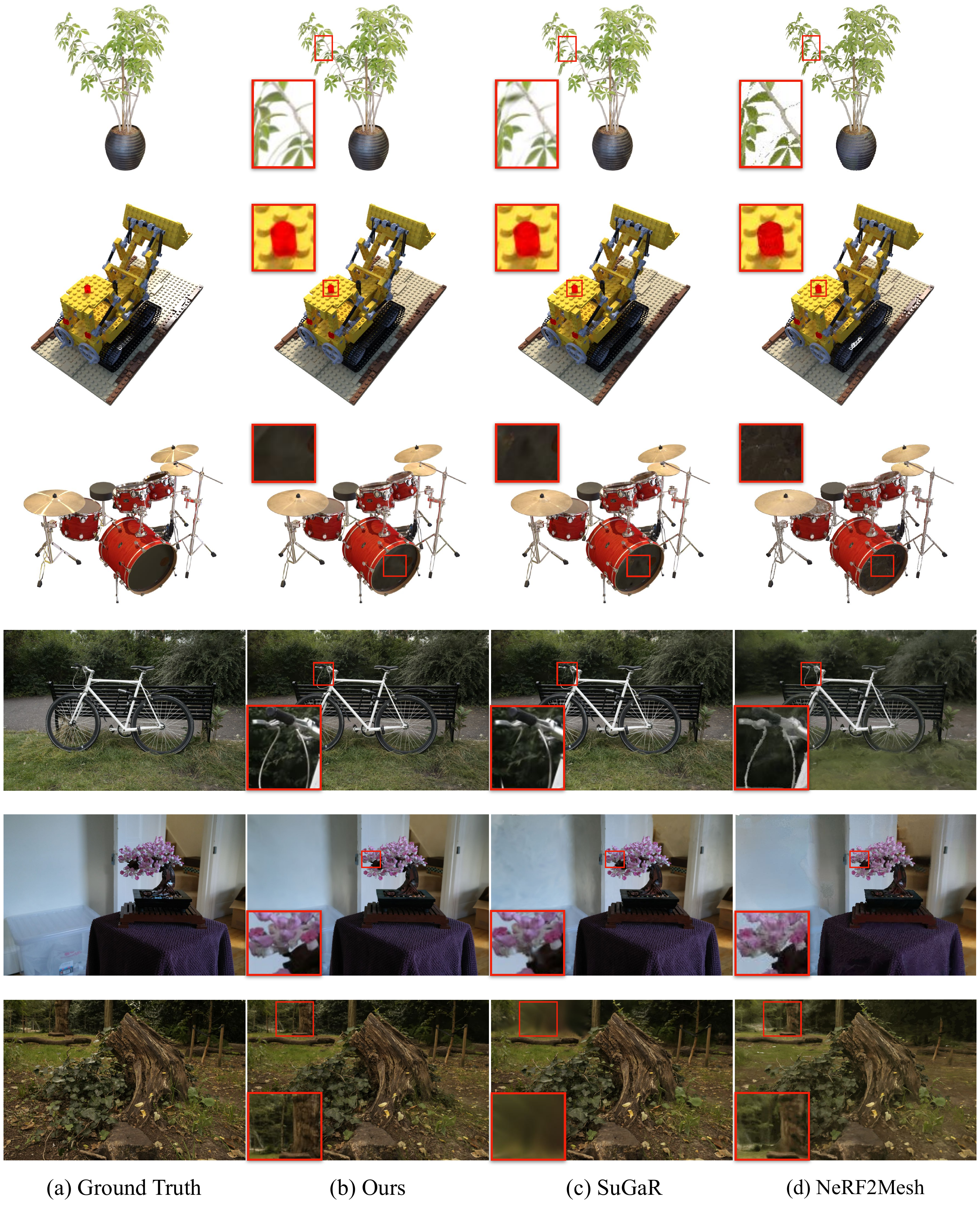}
  \caption{Qualitative comparisons with baseline methods (SuGaR \cite{Guedon23_SuGaR}, NeRF2Mesh \cite{Tang23_NeRF2Mesh}) on the NeRF-Synthetic \cite{Mildenhall20_NeRF} and Mip-NeRF360 datasets \cite{Barron22_Mip360}.}
  \label{fig:synthetic_mip_quality}
\end{figure*}

\subsection{Novel View Synthesis}

We evaluate our method on the novel view synthesis task. Quantitative results on the NeRF-Synthetic and Mip-NeRF360 datasets are shown in Tabs.~\ref{tab:NVS_syn} and \ref{tab:NVS_mip}, respectively. To ensure a fair comparison, we use a white background for the NeRF-Synthetic dataset. We compare our approach with recent methods, including both mesh-based and mesh-free approaches, and further distinguish whether the appearance representation is explicitly surface-bound.

On NeRF-Synthetic, our method achieves state-of-the-art quality among mesh-reconstruction methods and even surpasses some methods dedicated solely to view synthesis. Moreover, our performance is only marginally lower than the original 3DGS \cite{Kerbl23_3DGS}. On Mip-NeRF360, our method matches or exceeds the performance of state-of-the-art methods \cite{Guedon23_SuGaR, Tang23_NeRF2Mesh}. While recent Gaussian-based methods such as 2DGS \cite{Huang24_2DGS} and GOF \cite{Yu24_GOF} achieve slightly higher rendering metrics, they do not enforce explicit binding between appearance primitives and a reconstructed surface. Fig.~\ref{fig:synthetic_mip_quality} presents qualitative comparisons, showing that our method captures finer scene details than the baselines.

\begin{figure*}[tbp]
  \centering
  \includegraphics[width=1.0\textwidth]{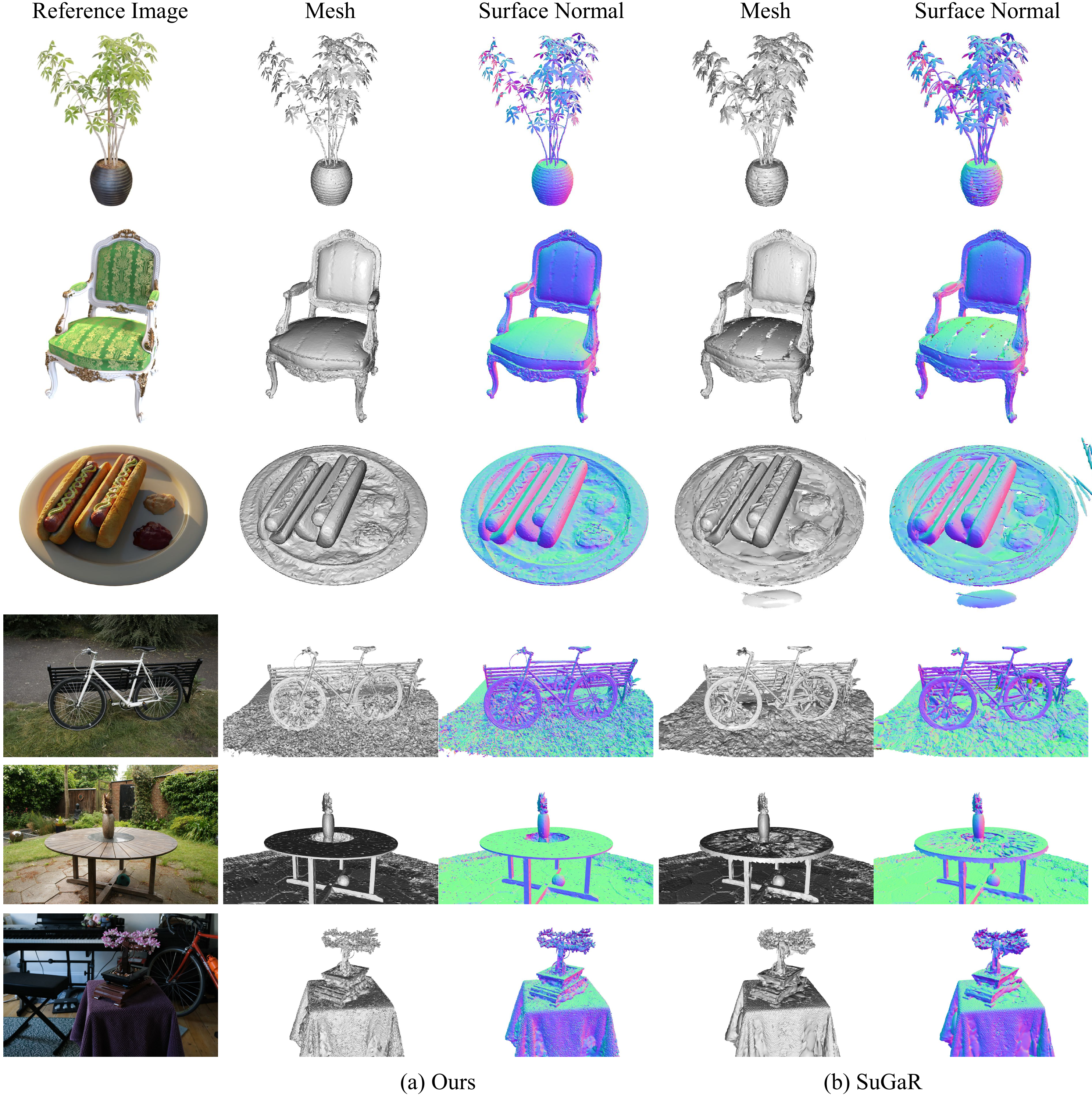}
  \caption{Qualitative comparisons of reconstructed meshes with SuGaR \cite{Guedon23_SuGaR} on the NeRF-Synthetic \cite{Mildenhall20_NeRF} and Mip-NeRF360 \cite{Barron22_Mip360} datasets.}
  \label{fig:im_mesh_normal}
\end{figure*}

\begin{table*}[tbp]
    \centering
    \caption{Quantitative evaluation of mesh quality and efficiency on the NeRF-Synthetic \cite{Mildenhall20_NeRF} dataset.}

    \begingroup
    \renewcommand{\arraystretch}{0.85}
    \setlength{\tabcolsep}{5pt}
    \begin{tabular}{l@{\quad}|@{\quad}c@{\quad}r@{\quad}|@{\quad}c|@{\quad}r}
    \toprule
    Method & CD ($10^{-3}$) & \#Faces & PSNR & Train Time \\
    \midrule
    NVdiffRec \cite{Munkberg22_Nvdiffrec}   & 8.15 & \cellcolor{first}\textbf{80k} & 29.05 & 52 mins \\
    NeRF2Mesh \cite{Tang23_NeRF2Mesh}       & \cellcolor{first}\textbf{5.06} & 192k & 31.04 & 46 mins \\
    SuGaR-15K \cite{Guedon23_SuGaR}         & 8.62 & 1000k& 32.09 & 103 mins \\
    Ours (w/o refine)                       & 8.50 & 141k & 27.68 & \cellcolor{first}\textbf{16 mins} \\
    Ours                                    & \cellcolor{second}7.27 & \cellcolor{second}100k & \cellcolor{first}\textbf{32.26} & \cellcolor{second}28 mins \\
    \bottomrule
    \end{tabular}
    \endgroup
    \\
    {\footnotesize `CD': Chamfer Distance between the ground-truth and reconstructed meshes.}
    \label{tab:chamfer_dist}
\end{table*}

\subsection{Surface Reconstruction}
While our primary goal is to efficiently learn a hybrid representation aligning Gaussians with mesh faces, the resulting surfaces are of high quality. In this section, we focus on comparisons with surface-bound methods that explicitly reconstruct a mesh and bind appearance to the surface, such as SuGaR \cite{Guedon23_SuGaR}. Fig.~\ref{fig:im_mesh_normal} shows the reconstructed meshes and their normal maps. Compared with the baseline, our method better preserves surface details, whereas SuGaR \cite{Guedon23_SuGaR} often produces floating patches and mesh holes.

For quantitative evaluation, we compute the Chamfer Distance (CD) between our reconstructed meshes and the ground-truth geometry from the NeRF-Synthetic dataset. The evaluation follows NeRF2Mesh \cite{Tang23_NeRF2Mesh}, in which 2.5M surface points are sampled from both the ground-truth and reconstructed meshes using ray casting. Tab.~\ref{tab:chamfer_dist} reports the CD values, face counts, rendering PSNR, and total training time for a comprehensive comparison. The results demonstrate that our reconstructed meshes achieve high geometric quality with significantly lower training cost. Furthermore, our Gaussian-based appearance model supports high-fidelity rendering even with fewer mesh faces compared to NeRF2Mesh.

Notably, the mesh obtained before refinement already exhibits satisfactory quality. For applications that require only an untextured mesh, our method delivers excellent results with just 16 minutes of training.

\begin{figure*}[tbp]
  \centering
  \includegraphics[width=0.72\textwidth]{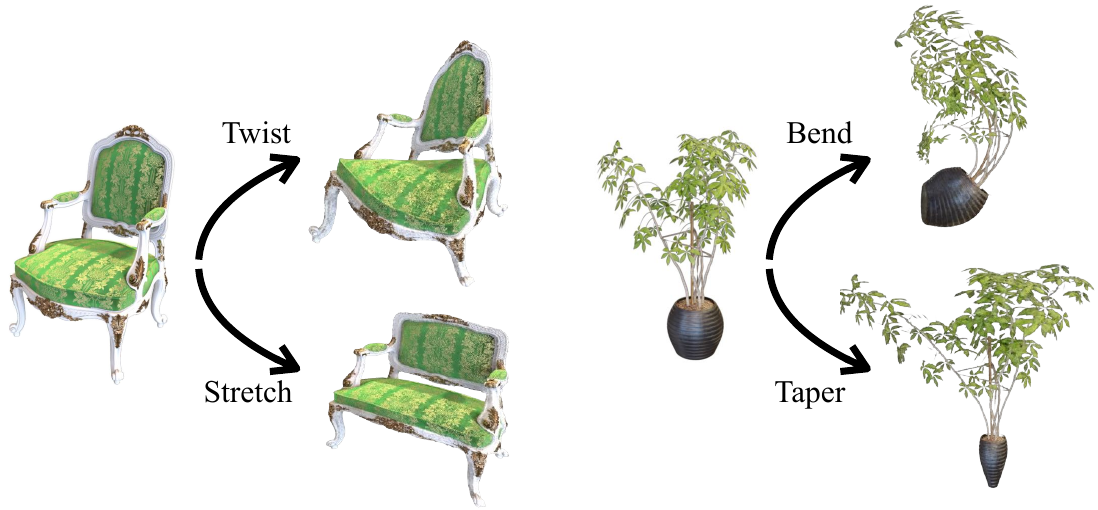}
  \caption{Object deformation using a hybrid representation of mesh and Gaussians. We employ four Blender modifiers to manipulate the mesh and subsequently render views using the bound Gaussians.}
  \label{fig:deform_mesh}
\end{figure*}

\subsection{Object Deformation by Hybrid Representation}
As discussed in the introduction, the hybrid representation not only enables efficient and high-quality rendering via Gaussians but also supports practical manipulation through the mesh, which benefits from well-established editing toolchains. To demonstrate the potential applications of our approach, we consider object deformation tasks. Specifically, we apply four standard mesh modifiers (Twist, Stretch, Bend, and Taper) in Blender to deform the learned mesh. The associated Gaussians are then updated using the binding method introduced in Sec.~\ref{sec:bind_gs}. The deformed object is rendered using Gaussian rasterization, as shown in Fig.~\ref{fig:deform_mesh}. More advanced deformation operations, such as those explored in \cite{Jiang24_VRGS, Gao24_MeshBasedGS}, are also compatible with our hybrid representation.

\subsection{Method Analysis}

\begin{table*}[tbp]
    \centering
    \caption{Efficiency evaluation. We report the training time and PSNR metric across different methods.}
    \begingroup
    \renewcommand{\arraystretch}{0.85}
    \setlength{\tabcolsep}{5pt}
    \begin{tabular}{lcccc}
    \toprule
    \multicolumn{1}{l}{\multirow{2}{1.1cm}{Method}} & \multicolumn{2}{c}{NeRF-Synthetic \cite{Mildenhall20_NeRF}} & \multicolumn{2}{c}{Mip-NeRF360 \cite{Barron22_Mip360}}  \\
    \cmidrule(lr){2-3} \cmidrule(lr){4-5} 
           & PSNR$\uparrow$ & Train Time$\downarrow$ & PSNR$\uparrow$ & Train Time$\downarrow$ \\
    \midrule
    NVdiffRec \cite{Munkberg22_Nvdiffrec}   & 29.05 & 52 mins & - & - \\
    NeRF2Mesh \cite{Tang23_NeRF2Mesh}       & 31.04 & 46 mins & - & - \\
    SuGaR-15K \cite{Guedon23_SuGaR}         & 32.09 & 103 mins & 27.27 & 132 mins\\
    Ours                                    & \textbf{32.26} & \textbf{28 mins} & \textbf{27.54} & \textbf{61 mins}\\
    \bottomrule
    \end{tabular}
    \endgroup
    \label{tab:efficiency_comparison}
    \vspace{3pt}
    \footnotesize
    \begin{minipage}{0.67\textwidth}
    `-' indicates that results for the method are not reported in its paper and that the method fails to reconstruct all scenes of the dataset using its published code.
    \end{minipage}
    
\end{table*}

\begin{table}[tbp]
    \caption{Ablation studies on adaptive covariance and the refinement stage. We report the average PSNR across all scenes of the NeRF-Synthetic \cite{Mildenhall20_NeRF} and Mip-NeRF360 \cite{Barron22_Mip360} datasets.}

    \centering
    \begingroup
    \renewcommand{\arraystretch}{0.85}
    \setlength{\tabcolsep}{5pt}
    \begin{tabular}{cc|cc}
    \toprule
    Adaptive Cov. & Refine & NeRF-Synthetic & Mip-NeRF360 \\
    \midrule
    
    $\text{\sffamily x}$  & $\checkmark$ & 32.02 & 27.19 \\
    $\checkmark$ & $\text{\sffamily x}$  & 27.78 & 24.44 \\
    $\checkmark$ & $\checkmark$ & \textbf{32.26} & \textbf{27.54} \\
    
    \bottomrule
    \end{tabular}
    \endgroup
    \label{tab:ablation}
\end{table}

\subsubsection{Efficiency}
We evaluate the efficiency of our method by measuring the average reconstruction time on the NeRF-Synthetic and Mip-NeRF360 datasets. All methods are evaluated under identical hardware settings for a fair comparison. As shown in Tab.~\ref{tab:efficiency_comparison}, our method not only achieves faster reconstruction but also delivers superior rendering quality.

\subsubsection{Ablation Studies}
\begin{center}
\begin{minipage}{\columnwidth}
    \centering
    \captionof{table}{Ablation study on the number of Gaussians per face. This table shows the PSNR results on the Mip-NeRF360 \cite{Barron22_Mip360} dataset.}
    \label{tab:ablation_gs_perface}
    \vspace{1pt}
    \includegraphics[width=0.7\linewidth]{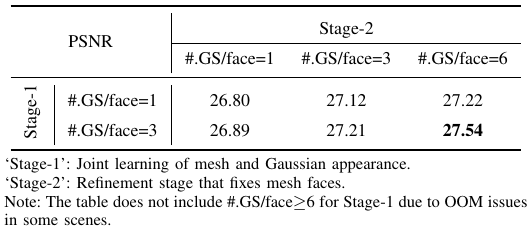}
\end{minipage}
\end{center}

We analyze the impact of three design factors: (1) adaptive Gaussian covariance, (2) the refinement stage, and (3) the number of Gaussians per face. 

To isolate the effect of adaptive Gaussian covariance, we replace $\boldsymbol{M}$ in \eqref{eq:cov_final} with the identity matrix, effectively removing the shape-dependent transformation. As shown in Tab.~\ref{tab:ablation}, both adaptive covariance and the refinement stage lead to significant improvements in rendering quality.

The number of Gaussians per face presents a trade-off between computational cost and rendering performance. As shown in Tab.~\ref{tab:ablation_gs_perface}, increasing the number of Gaussians per face improves rendering quality. To balance rendering quality with memory usage and training time, we use 3 Gaussians per face during the joint learning stage. For the refinement stage, we increase this number to 6 to further enhance surface fidelity.

\FloatBarrier

\section{Conclusion}

We have presented an end-to-end method that jointly learns explicit triangle-mesh geometry and surface-bound 3D Gaussian appearance from multi-view images. Differentiable 3DGS rendering directly supervises both components through the photometric loss, avoiding a separate appearance-reconstruction stage. Experiments demonstrate high-quality rendering and surface reconstruction, efficient training, adaptation to local scene updates, and mesh-based manipulation.

\appendix

\section{Additional Results and Details}

\begin{table*}[tbp]
    \centering
    \caption{Per-scene quantitative results on the Mip-NeRF360 dataset \cite{Barron22_Mip360}.}
    \begingroup
    \renewcommand{\arraystretch}{0.85}
    \setlength{\tabcolsep}{4pt}
    \begin{tabular}{c@{\quad}|@{\quad}c@{\quad}c@{\quad}c@{\quad}c@{\quad}c@{\quad}c@{\quad}c@{\quad}|@{\quad}c}
        \toprule
        Metric  & Bicycle & Bonsai & Counter & Garden & Kitchen & Room & Stump & Mean \\
        \midrule						
        PSNR    & 24.18 & 31.47 & 27.16 & 26.41 & 28.96 & 29.71 & 24.91 & 27.54 \\
        SSIM    & 0.722 & 0.945 & 0.888 & 0.829 & 0.898 & 0.908 & 0.712 & 0.843 \\
        LPIPS   & 0.260 & 0.175 & 0.212 & 0.161 & 0.162 & 0.221 & 0.291 & 0.212 \\
        \bottomrule
    \end{tabular}
    \endgroup
    \label{tab:per_scene_mip}
\end{table*}

\subsection{Details for SDF Grid Optimization}
This section includes the methods and configurations used to mitigate the storage and computational demands of the SDF grid:

\paragraph{View-Dependent Grid Optimization.}
For the given grid nodes $\{\boldsymbol{x_n}\}_{n=1}^N$, the visible subset is calculated by:
\begin{align}
\{\boldsymbol{\hat{x}_m}\} := \left\{\boldsymbol{x_n} \big| \text{NDC}(\boldsymbol{x_n})\in[-1,1]^3 \right\},
\end{align}
where $\text{NDC}(\boldsymbol{x_n})$ transforms the world-space coordinates $\boldsymbol{x_n}$ to normalized device coordinates (NDC) via the projection matrix. The marching algorithm is only applied to cells whose eight corner nodes are within $\{\boldsymbol{\hat{x}_m}\}$.

\paragraph{Progressive Grid Refinement.}
Given a bounding box (BBox) with size ($L_x$, $L_y$, $L_z$) that tightly encloses the foreground region $U$, and an expected cell count $C$, the cubic cell size of the SDF grid is set to $(L_x/s) \times (L_y/s) \times (L_z/s)$, where $s=\sqrt[3]{L_x \cdot L_y \cdot L_z / C}$. During optimization, we perform 4 coarse-to-fine operations, incrementally increasing the cell count by $1.5\times$ each step until the target count $C$ is reached.

\subsection{Configurations of Face Gaussians}
As discussed in Sec. \ref{sec:bind_gs}, each face is associated with $K$ Gaussians. The choice of $K$ influences the barycentric coordinates of the Gaussian centers and the radius $r$ employed in the covariance matrix $\boldsymbol\Sigma_e$ in Eq. \eqref{eq:cov_final}. For $K=3$, we use 
\begin{align}
\boldsymbol{\xi}_1 &=
\left[(3-\sqrt{3})/6, (3-\sqrt{3})/6, \sqrt{3}/3\right], \notag\\
\boldsymbol{\xi}_2 &=
\left[(3-\sqrt{3})/6, \sqrt{3}/3, (3-\sqrt{3})/6\right], \notag\\
\boldsymbol{\xi}_3 &=
\left[\sqrt{3}/3, (3-\sqrt{3})/6, (3-\sqrt{3})/6\right], \notag
\end{align}
and radius $r=\Vert \boldsymbol v_1 - \boldsymbol v_2 \Vert/(2\sqrt{3}+2)$. For $K=6$, we use 
\begin{align}
\boldsymbol{\xi}_1 &=
\left[(3-2\sqrt{3})/6, (3-2\sqrt{3})/6, 2\sqrt{3}/3\right], \notag\\
\boldsymbol{\xi}_2 &=
\left[(3+2\sqrt{3})/12, (3-\sqrt{3})/6, (3+2\sqrt{3})/12\right], \notag\\
\boldsymbol{\xi}_3 &=
\left[(3-\sqrt{3})/6, (3+2\sqrt{3})/12, (3+2\sqrt{3})/12\right], \notag\\
\boldsymbol{\xi}_4 &=
\left[2\sqrt{3}/3, (3-2\sqrt{3})/6, (3-2\sqrt{3})/6\right], \notag\\
\boldsymbol{\xi}_5 &=
\left[(3+2\sqrt{3})/12, (3+2\sqrt{3})/12, (3-\sqrt{3})/6\right], \notag\\
\boldsymbol{\xi}_6 &=
\left[(3-2\sqrt{3})/6, 2\sqrt{3}/3, (3-2\sqrt{3})/6\right], \notag
\end{align}
and radius $r=\Vert \boldsymbol v_1 - \boldsymbol v_2 \Vert/(2\sqrt{3}+4)$.

\subsection{Derivation of Adaptive Gaussian Covariance}
\begin{figure}[tbp]
  \centering
  \includegraphics[width=0.26\textwidth]{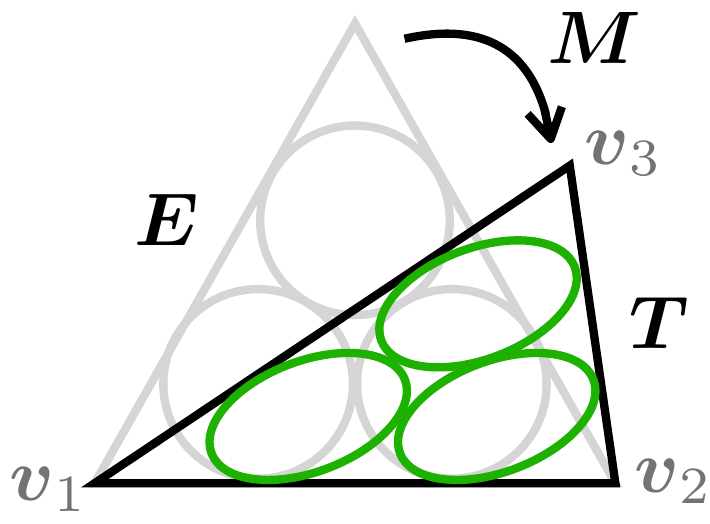}
  \caption{The linear transformation applied to Gaussians for an irregular triangle.}
  \label{fig:E_to_T}
\end{figure}
This section presents how to obtain the matrix $\boldsymbol M$ used in \eqref{eq:cov_final} for a target triangle $(\boldsymbol{v}_1, \boldsymbol{v}_2, \boldsymbol{v}_3)$.

As shown in Fig.~\ref{fig:E_to_T}, the linear transformation applied to a reference equilateral triangle will result in the target triangle. In the local coordinate system defined in \eqref{eq:triangle_coord}, the target triangle is denoted as $\boldsymbol{T}=[\boldsymbol{v}'_1, \boldsymbol{v}'_2, \boldsymbol{v}'_3]$, and the reference equilateral triangle is denoted as $\boldsymbol{E}=[\boldsymbol{v}'_1, \boldsymbol{v}'_2, \hat{\boldsymbol{v}}'_3]$, where $\hat{\boldsymbol{v}}'_3=l \cdot [0, \frac{1}{2}, \frac{\sqrt{3}}{2}]^T$, with $l=\Vert\boldsymbol{v}_2-\boldsymbol{v}_1\Vert$.

By solving the equation $\boldsymbol M\boldsymbol E = \boldsymbol T$, we obtain:
\begin{align}
\boldsymbol M = 
\begin{bmatrix}
1 & 0 & 0 \\
0 & 1 & \frac{2\boldsymbol v'^{(2)}_3-l}{\sqrt{3} l} \\
0 & 0 & \frac{2\boldsymbol v'^{(3)}_3}{\sqrt{3} l} \\
\end{bmatrix}
,
\end{align}
where $\boldsymbol v'^{(i)}_3$ denotes the $i$-th component of $\boldsymbol{v}'_3$.

\subsection{More Results}
Tab.~\ref{tab:per_scene_mip} includes the per-scene results on the Mip-NeRF360 dataset \cite{Barron22_Mip360}.



\FloatBarrier

\section*{CRediT authorship contribution statement}
\noindent\textbf{Ancheng Lin:} Conceptualization, Methodology, Software, Validation, Formal analysis, Investigation, Visualization, Writing -- Original Draft. \textbf{Tianqing Su:} Methodology, Software, Validation, Investigation, Visualization, Writing -- Original Draft. \textbf{Zuo Yuan:} Methodology, Validation, Writing -- Review \& Editing. \textbf{Quanke Su:} Methodology, Validation, Writing -- Review \& Editing. \textbf{Samuel S. Mao:} Supervision, Writing -- Review \& Editing. \textbf{Yusheng Xiang:} Conceptualization, Supervision, Project administration, Writing -- Review \& Editing.

\section*{Funding}
This work was supported by the Jiangsu Provincial Department of Industry and Information Technology under Project No. JSSCRC20250022.

\section*{Declaration of competing interests}
The authors declare that they have no known competing financial interests or personal relationships that could have appeared to influence the work reported in this paper.

\section*{Data availability}
The datasets analyzed in this study are publicly available from the following repositories: NeRF-Synthetic, \url{https://github.com/bmild/nerf}; and Mip-NeRF360, \url{https://jonbarron.info/mipnerf360}. The source code for the proposed method is available at \url{https://github.com/Cenbylin/DMGS}.

\section*{Declaration of generative AI and AI-assisted technologies in the manuscript preparation process}
During the preparation of this work, the authors used ChatGPT solely to improve the language and readability of the manuscript. After using this tool, the authors reviewed and edited the content as needed and take full responsibility for the content of the publication.

\setlength{\bibsep}{0pt}
\bibliographystyle{elsarticle-num-names}
\bibliography{biblo}

\end{document}